\documentclass[conference]{IEEEtran}

\usepackage{graphicx}
\usepackage{amsmath}
\usepackage{amssymb}
\usepackage{booktabs}
\usepackage{tablefootnote}

\usepackage[pagebackref,breaklinks,colorlinks]{hyperref}

\usepackage[capitalize]{cleveref}
\crefname{section}{Sec.}{Secs.}
\Crefname{section}{Section}{Sections}
\Crefname{table}{Table}{Tables}
\crefname{table}{Tab.}{Tabs.}

\begin{document}


\title{Effective Utilisation of Multiple Open-Source Datasets to Improve Generalisation Performance of Point Cloud Segmentation Models}

\author{Matthew Howe$^{1}$, Boris Repasky$^{1,2}$, Timothy Payne$^{2}$\\
Australian Institute for Machine Learning, The University of Adelaide $^{1}$\\
Lockheed Martin Australia STELaRLab $^{2}$\\
{\tt\small matthew.howe@adelaide.edu.au, boris.repasky@adelaide.edu.au, timothy.m.payne@lmco.com}
}
\maketitle

\begin{abstract}
Semantic segmentation of aerial point cloud data can be utilised to differentiate which points belong to classes such as ground, buildings, or vegetation.
Point clouds generated from aerial sensors mounted to drones or planes can utilise LIDAR sensors or cameras along with photogrammetry.
Each method of data collection contains unique characteristics which can be learnt independently with state-of-the-art point cloud segmentation models.
Utilising a single point cloud segmentation model can be desirable in situations where point cloud sensors, quality, and structures can change.
In these situations it is desirable that the segmentation model can handle these variations with predictable and consistent results.
Although deep learning can segment point clouds accurately it often suffers in generalisation, adapting poorly to data which is different than the training data. 
To address this issue, we propose to utilise multiple available open source fully annotated datasets to train and test models that are better able to generalise. 
In this paper we discuss the combination of these datasets into a simple training set and challenging test set.
Combining datasets allows us to evaluate generalisation performance on known variations in the point cloud data.
We show that a naive combination of datasets produces a model with improved generalisation performance as expected.
We go on to show that an improved sampling strategy which decreases sampling variations increases the generalisation performance substantially on top of this.
Experiments to find which sample variations give this performance boost found that consistent densities are the most important.
\end{abstract}

\IEEEpeerreviewmaketitle
\section{Introduction}\label{sec:intro}
Semantic segmentation of aerial point clouds into their respective object classes is a challenging task \cite{survey}.
Relationships must be found between points to extract and represent local structure and wider context of the scene to accurately classify the class.
For example a local flat plane structure could be part of the ground or a building roof depending on the wider elevation context.
Current research focusing on point cloud segmentation has resulted in capable models to perform the task \cite{pointnet, pointnetpp, randlanet}.
These models have been shown to perform well when trained and tested on single source datasets.

Common variations contained in point cloud data include the type of sensor, topography, building architecture, and the overall quality, density, and noise contained \cite{SUM, DALES, swiss, ISPRS}.
These variations are often out of the training domain and hence are detrimental to model performance \cite{2018arXiv180601759H, 2022arXiv220712654Y, coralgeneralisation, classificationgeneralisation}.
In some situations these variations are unknown at inference time.
In this situations it is desirable to have a singular model which has the ability to perform well, albeit possibly worse than a specialised model, despite the variations.


While others have focused on achieving state-of-the-art results by training and testing on the same dataset \cite{KPConv, shellnet, pointnet, pointnetpp, randlanet}, we focus on improving the generalisation performance of a commonly used point cloud model using limited training examples.
Improving generalisation performance aims to achieve more predictable and higher performance on out of domain data.
To achieve this we utilise multiple open source datasets collected from aerial sensors of outdoor scenes.
The datasets contain differences in the data collection method, scale, quality, local geography, and building architecture.
We combine multiple open source datasets into a cohesive dataset with a small, simple training and validation set and a challenging test set.
This dataset split allows us to test the generalisation performance of the model with known out of domain data.


In this paper we first present a comprehensive and challenging benchmark dataset created from multiple open source datasets.
We provide our justification and methodology of how datasets were selected and combined.
We show a naive approach to training a model on the combined dataset is effective and produces improved results.
Our improved sampling method although simple is effective in increasing the performance further. 
To understand the factors influencing generalisation performance we present an ablation of the important factors of model performance on the combined test set.

\section{Related work}\label{sec:formatting}
In this section we describe the open source labelled aerial datasets utilised in this paper and discuss previous point cloud segmentation methods related to our work.

\subsection{Aerial datasets}
Table \ref{tab:dataset_summary} contains a summary of each identified dataset. 
The datasets are captured using aerial sensors typically mounted to drones and aircraft. 
After data collection laser scans or multiple images of the scene are compiled into point clouds which span large areas.
Due to the size of the point clouds they are often broken into smaller tiles.
We note the year of data collection, the sensor type used, the number of classes and the average overhead point density.

The ISPRS dataset \cite{ISPRS} is the oldest of this type of dataset being collected in 2012. 
ISPRS is relatively sparse and small dataset in comparison to the newer datasets. 
SUM, Dublin city, and SenSatUran \cite{SUM, dublin, sensat} datasets contain point clouds from single city urban areas which have been tiled to cover large areas. 
Swiss3D cities \cite{swiss} contains point clouds from three cities in Switzerland which have notably different architecture, terrain, and city layouts. 
Swiss3D cities also includes varying density point clouds of the same area. 
The H3D \cite{H3D} dataset contains both LIDAR and photogrammetry data of the same area over a number of years resulting in variable quality of dataset. 

Classes for these datasets contain ground, vegetation, vehicles, power lines, urban furniture, fences, and buildings. 
Some datasets further differentiate classes, for example splitting vegetation into trees or shrubs and buildings into their component parts such as walls, roof, and windows.

There are datasets of aerial point clouds which we do not consider or run experiments on including LASDU, 3DOMCity, and Bordeaux \cite{LASDU,3DOMcity,Bordeaux}. 
The 3DOMCity dataset was made utilising a small model of a city block and hence does not cover an extensive area.
Utilisation of this dataset could lead to artificially inflated generalisation results due to the smaller nature.
Both the LASDU and Bordeaux datasets were not accessible at the time of evaluation and are also excluded from our evaluations.

\begin{table*}
    \centering
    \caption{A summary of publicly available labelled 3D LIDAR and photogrammetry datasets}
    \resizebox{\textwidth}{!}{
        \begin{tabular}{c||c|c|c|c}
        \hline
        \textbf{Name} & \textbf{Year} & \textbf{Sensor Type} & \textbf{Classes} & \textbf{Point density $p/m^2$} \\
        \hline
        \hline
        ISPRS-Vaihingen \cite{ISPRS} & 2012 & LIDAR   & 9 & 1.52 \\
        \hline
        H3D \cite{H3D} & 2016, 2018.3, 2018.11, 2019 & LIDAR,Photogrammetry  & 11& 2016.3: 10.31 \& 2018.3: 1025.32 \& 2018.11: 1105.32 \&  2019.3: 1454.12  \\
        \hline
        Dublin city \cite{dublin} & 2019 & LIDAR  &  13 & 105.25 \\
        \hline
        DALES \cite{DALES} & 2020 & LIDAR & 8 & 42.83 \\
        \hline
        LASDU \cite{LASDU} & 2020 & LIDAR & 5 & 3.45 \\
        \hline
        Bordeaux \cite{Bordeaux} & 2020 & LIDAR, Photogrammetry & 5 & 30.21 \\
        \hline
        Swiss3Dcities \cite{swiss} & 2021 & Photogrammetry  & 5 & Sparse:1.75 \& Medium: 65.343\\
        \hline
        SenSatUrban \cite{sensat} & 2021 & Photogrammetry  & 13 & 261.54\\
        \hline
        SUM \cite{SUM} & 2021 & Photogrammetry & 6 & 30: 51.07 \& 300: 458.13\\
        \hline

        \end{tabular}
    }

    \label{tab:dataset_summary}
\end{table*}

The datasets shown in table \ref{tab:dataset_summary} were all created to aid in the training of segmentation models.
Despite this there is considerable variation in the motivation and execution of each dataset collection which results in each having unique characteristics.
For example some datasets focus on a particular area such as DALES while others focus on collecting data with as much variation as possible like Swiss3DCitites.
These variations allow us to quantify the models ability to generalise across these varying characteristics.



\subsection{LIDAR Segmentation}\label{sec:segmentationmethods}
This section provides a brief overview of the point cloud segmentation models utilised in our work.
In this paper we utilise point-wise multi-layer perceptron methods for point cloud segmentation as overall performance is reasonably high \cite{survey}.
Other methods utilising convolutions \cite{KPConv, pointCNN, convpoint}, recurrent neural networks \cite{3Prnn}, and graphs \cite{DGCNN, SPG, sspspg} were not considered for this particular work.

Point-wise multi-layer-perceptron (MLP) methods such as PointNet \cite{pointnet}, PointNet++ \cite{pointnetpp} and RandLA-Net \cite{randlanet} use a MLP to create point wise feature representations for the surrounding points. 
Specifically, PointNet predicts the local point features for each point, the local features are aggregated to a global feature, and finally in the case of semantic segmentation the local and global features are concatenated and passed through a MLP to produce point-wise segmentation labels. 
PointNet++ built on this work using hierarchical neural networks in order to extract the local features better than PointNet, this involves the addition of sampling and clustering on the point clouds. 
The sampling and clustering is computationally expensive and is addressed in the RandLA-Net paper. 
RandLA-Net uses random sampling instead of point selection approaches which attempt to find local region centroids. 
As this process inevitably loses local features they add the local spatial encoding (LocSE) module which increases the perceptive field of each point in order to overcome this. RandLA-Net has higher performance on most benchmarks with less computation than PointNet++ and PointNet \cite{survey}.


\subsection{Generalisation for aerial point cloud segmentation}
Training and testing point cloud segmentation models on multiple datasets has been researched in the past.
The authors of \cite{classificationgeneralisation} use three dataset to test model transfer, DALES, ISPRS, and Bordeaux. 
In this case the generalisation performance is measured by training on either DALES or ISPRS and tested on ISPRS or Bordeaux.
This train and test split does not provide enough variability and diversity within the test set to identify trends for generalisation.

The authors of \cite{coralgeneralisation} utilise two LIDAR datasets: ISPRS and LASDU.
Models are trained on one dataset and tested on another using a variety of data augmentations and a popular method of domain generalisation, deep CORAL \cite{deepcoral}.
Generalisation performance is measured by comparing the model transfer performance on each respective test set.
This combination of datasets tests how model performance transfers between datasets which are collected with a similar method on different cities. 

Although these training and test splits provide sufficient signal for model transfer performance they do not contain enough diversity to evaluate overall generalisation performance for our task.
In this paper we aim to significantly increasing the amount and diversity of data that generalisation performance is evaluated on and provide characteristics which can be used to observe performance on sub-problems.

\section{Data preparation}
To train a single network across multiple datasets each dataset must be homogenised to a single format. 
Each point cloud dataset has differences in the format the data is stored in, the contents of the dataset such as class labels, and data channels from the sensors. 
In this section we discuss our methodology in the combination of the datasets to enable the generalisation evaluations.

\subsection{Dataset selection}
We aim to select a train and test set that enables us to approximate and evaluate the generalisation performance of a particular method. 
We believe that separating a dataset into a simple training set and a more complicated, challenging, and diverse test set allows us to draw such conclusions.
Evaluating the complexity of a particular dataset is based on qualitative assessment of the contents and evaluations they may enable.
We assess each dataset on content diversity, data collection method, and quality to determine if the dataset forms a challenging benchmark in comparison to others. 
Although we could choose a large training set by splitting the available datasets in half we choose to maximise the number of different testing sets.
We believe that this will provide the strongest signal of generalisation performance and decrease the chances that performance increases are dependent on dataset.

Note: When considering low, medium, and high density point clouds we refer to the ranges of $pts/m^2 < 2$, $5 < pts/m^2 < 100$, $100 < pts/m^2$, respectively.

\begin{figure*}
    \centering
    \resizebox{\textwidth}{!}{
        \begin{tabular}{c}
            \includegraphics[width=0.5\textwidth]{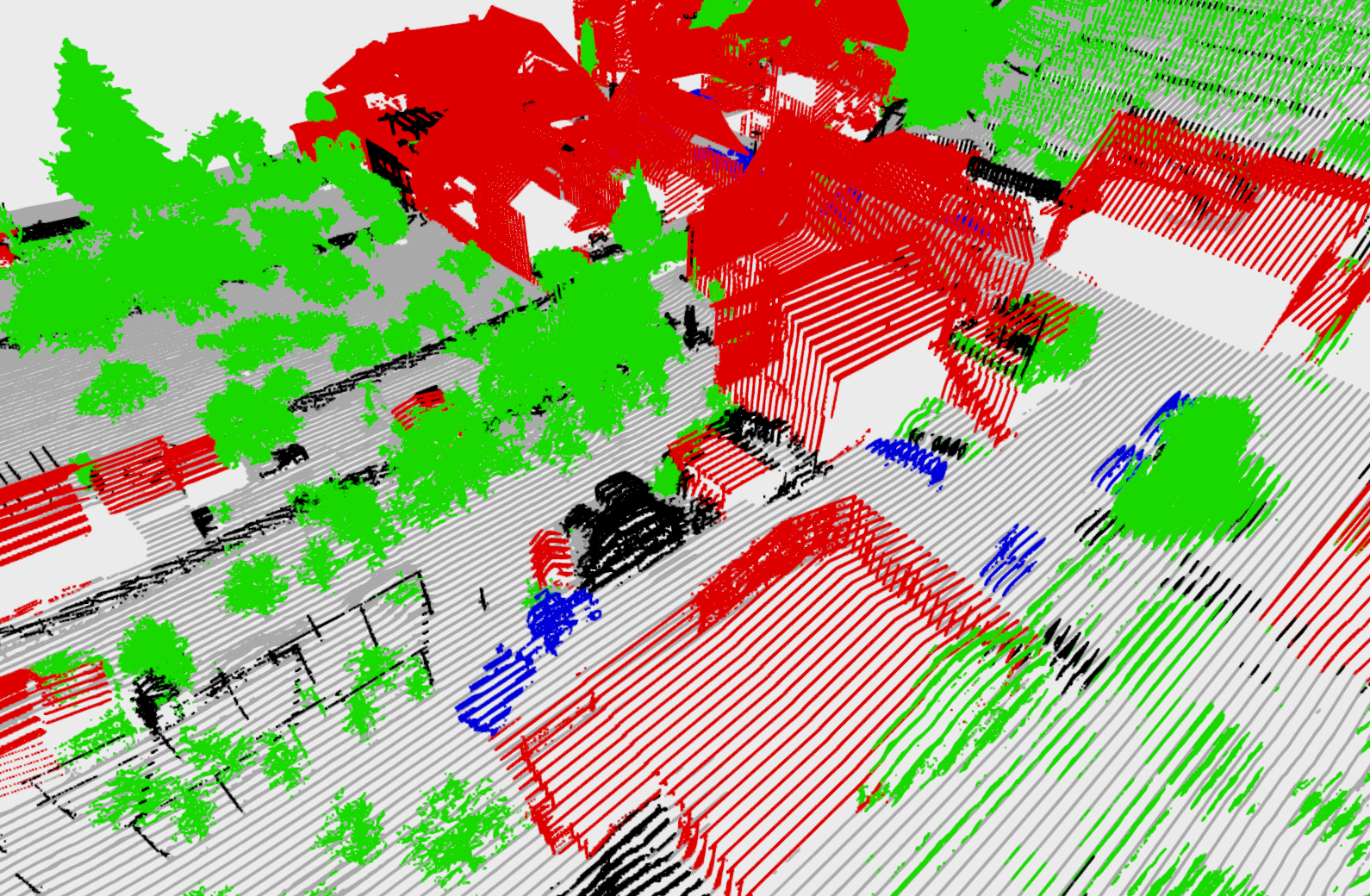}
            \includegraphics[width=0.5\textwidth]{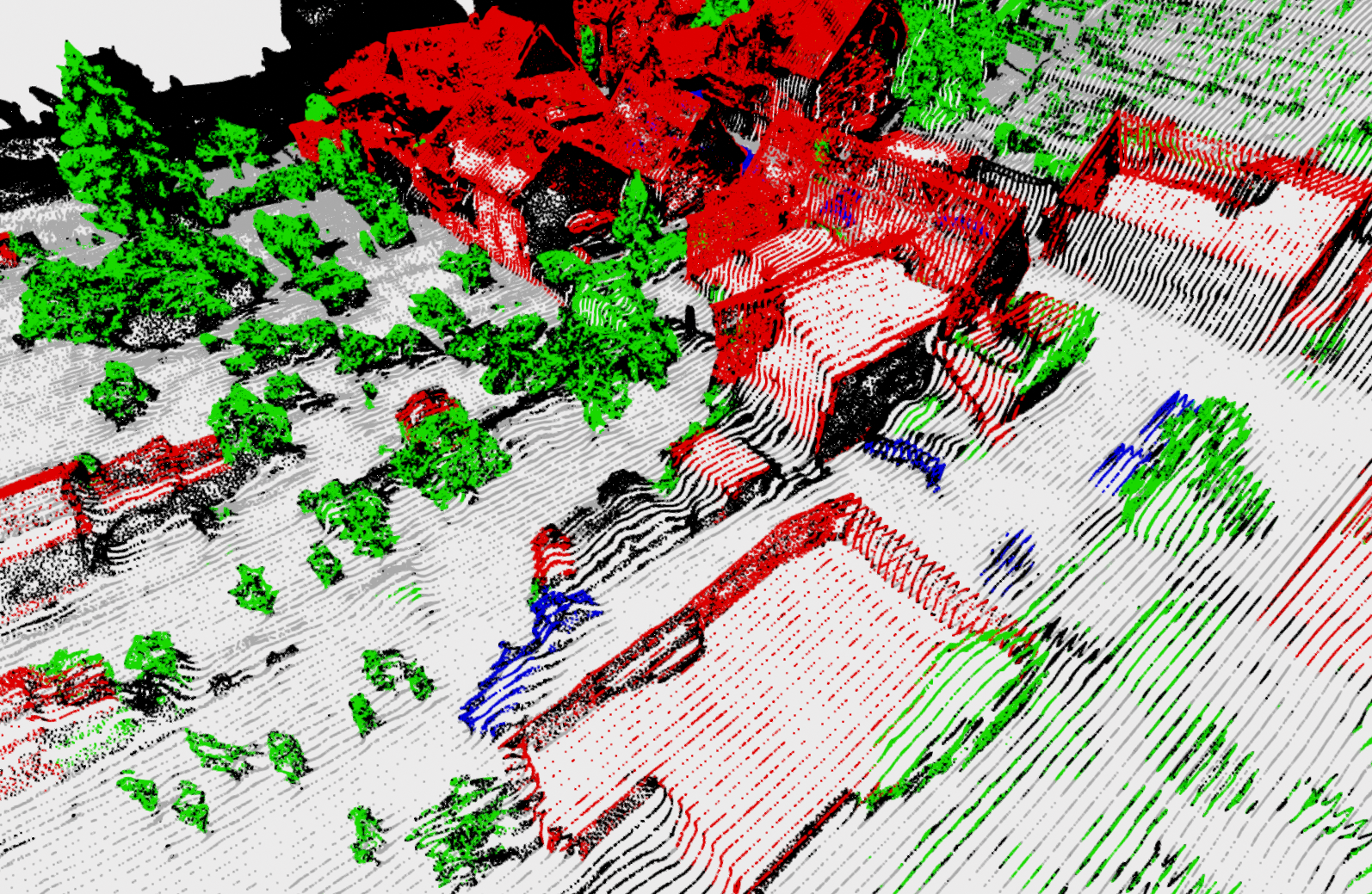}
        \end{tabular}
    }
    \caption{Comparison between the Mar2019 LIDAR and photogrammetry data qualities. The authors of H3D transfer labels form the LIDAR generated point cloud to the photogrammetry data and hence some points are labelled undefined (black).}
    \label{fig:h3dmeshLIDARcomparison}
\end{figure*}

\subsubsection{Test set}
Our selected test set consists of five datasets including Swiss3DCities, H3D, Sensat, DublinCity, and ISPRS.
This combination includes large amount of diversity with densities ranging from $1.7pts/m^2$ to $1454pts/m^2$, sensor types used, contents of the scene, and quality of the dataset.

The Swiss3DCities dataset contains point clouds of three cities Zurich, Davos, and Zug each of which have different urban densities and topography. 
For example Zug contains a rural town with spaced out buildings, Davos contains point clouds with large elevation changes, and Zurich is the dense urban centre.
Additionally we can compare performance on the low and medium density point clouds provided for the same regions.

The H3D dataset contains data over a smaller area than Swiss3DCities but contains data from multiple years as well as photogrammetry mesh and LIDAR point clouds of the same area.
This should allow us to make comparisons between model performance on LIDAR vs photogrammetry datasets by taking out factors such as content.
The photogrammetry meshes provided in the H3D dataset are not of the same quality as the LIDAR point clouds.
For example the 2019 mesh point cloud contains 91\% less points compared to the LIDAR counterpart.
Although the photogrammetry point clouds were generated in conjunction with the LIDAR data these point clouds have much lower quality and detail, see Figure \ref{fig:h3dmeshLIDARcomparison}.
This dataset allows for comparisons of the model to perform with data collected at different times by different sensors and with different qualities.

The DublinCity, ISPRS, and Sensat datasets all contain data from a single urban area.
The ISPRS dataset is utilised as a test set since it is not large enough to be included as a training set and can provide a measure of generalisation to older point cloud data collection.
DublinCity offers a large dataset collected using LIDAR which frequently penetrates through buildings, a phenomenon that was not observed in DALES and H3D as frequently. 
The Sensat dataset contains data collected using photogrammetry with high detail and covers a diverse cityscape with commercial, industrial, and residential areas.
These three datasets offer additional signal to support the generalisation characteristics which may be observed utilising the H3D and Swiss3DCities datasets.

There are some point clouds within the datasets which are not suitable for being combined into the test set and hence are removed or not evaluated.
The March 2018 LIDAR and photogrammetry data of H3D dataset does not contain the full area covered by the March 2016, November 2018, and March 2019.
Evaluating on this set results in much higher performance as it is a smaller simpler point cloud.
The March 2016 point clouds do not include photogrammetry data and is hence also excluded from evaluation.

\paragraph{Training and validation set}
For the training set we select two datasets, DALES and SUM.
These datasets do contain a satisfactory amount of variation which should allow a model to learn the fundamentals to perform well on the test set.
The data contains dense urban, rural, and residential areas as well as data collected from both LIDAR and photogrammetry.

Using this combination of test and train set we can test generalisation performance on variables such as sensor type, variations in quality, data collection times and methods.
The test set also contains data which is not present in the training set like lower quality and low density point clouds, different content, and more elevation variation.
This should allow for evaluation of the weaknesses and strengths of a segmentation model for generalising across variable characteristics which are not in the training set.
For the validation set we utilise the same train/val split given in the DALES and SUM papers.

\subsection{Class combinations}
Each dataset included in Table \ref{tab:dataset_summary} includes different classes, categories, and detail in the labels. 
In order to combine and train a single model on all datasets these classes need to be combined. 
In Table \ref{tab:datasetclasses} we show that each dataset's classes can be condensed into three larger categories: ground, buildings, vegetation. 
Each column shows a dataset and the classes with the label name from their respective paper that have been combined into the rows category.
Categories are formed when there are matching classes or sub-classes between all datasets.
Street furniture is not labelled accordingly in every dataset and hence does not form a category while buildings are labelled in every dataset.
Any classes which remain unmatched into a larger category have labels overwritten as \textit{undefined}.

In combining the datasets compromises must be made to form the unified set. 
Despite some datasets having highly detailed labels the resulting combined dataset will contain relatively few unique labels. 
We believe that, although few in number, this class combination still present a challenging task when coupled with differences between the dataset contents.
In this work we train models utilising the three classes present in Table \ref{tab:datasetclasses} to train and test generalisation performance.

\begin{table*}
    \centering
    \caption{Shared class categories that could be used for generalisation training across multiple open source datasets}
    \resizebox{\textwidth}{!}{
        \begin{tabular}{c||c|c|c|c|c|c|c}
        \hline
        \textbf{Class Category} & \textbf{DALES} & \textbf{H3D} & \textbf{Sensat} & \textbf{DublinCity} & \textbf{SUM} & \textbf{Swiss3Dcities} & \textbf{ISPRS 3D Benchmark} \\
        \hline
        \hline
        Ground-like & Ground & \shortstack{Low vegetation\\Soil/Gravel\\ Impervious surface} & \shortstack{Ground \\Bridge \\Parking \\Traffic road \\Footpath} & Ground & Terrain & Terrain & \shortstack{Low vegetation \\Impervious surface} \\
        \hline
        Building-like & Buildings & \shortstack{Roof \\ Facade \\ Chimney} & Building & Building & Building & Building & \shortstack{Roof \\ Facade}\\
        \hline
        Vegetation-like & Vegetation & Shrub & Vegetation & Vegetation & High Vegetation & Vegetation & \shortstack{Shrub \\ Tree} \\ 
        \hline
        \end{tabular}
    }
    \label{tab:datasetclasses}
\end{table*}

\subsection{Dataset formatting}\label{sec:DatasetFormatting}
After obtaining all of the datasets we convert them into a format which is consistent to load and process. 
This involves making all the data the same format, re-issuing the labels for each point and saving them independently.
Meshes were transformed into point clouds using open source software \cite{cloudcompare}.
The final data structure contains a dictionary with points, intensity, rgb, labels, and a KD-Tree for each tile contained in the dataset.
The KD-Tree is used to sample the point cloud at training and testing time.

\section{Model training and evaluation}\label{sec:training}
This section covers our training and testing methodology including the model used, important parameters, and training regime.
Our aim in training and testing the models is not to optimise a particular training strategy but to measure the effects of different training methods on the generalisation performance.
For this reason we use a simple training methodology.
In these experiments we do not utilise RGB and intensity data to simplify the training methodology and comparisons.

In our experimentation we use an off the shelf RandLA-Net model for it's relatively high performance on aerial point cloud segmentation tasks.
Our specific implementation uses 5 layers with sub-sampling ratios of 4, 4, 4, 4, and 2, respectively.
We use 16 nearest-neighbours sampling for the LocSE module. 
Since we do not utilise any additional channels such as RGB or intensity the input dimensions are simply $3 \times N$ for the location of the points and the output dimension is $3\times N$, for each class considered.
Unless otherwise stated the model is trained on samples of $65536$ points.

Each time the model is trained for a maximum of 100 epochs of 500 samples with a batch size of 12. 
Each training dataset is sampled equally, in our case 250 samples for each dataset per epoch.
Classes are weighted by the inverse occurrence in the training data which is approximately $[0.2, 0.4, 0.4]$ for ground, vegetation, and building, respectively.
We use a learning rate of $0.01$ and reduce it when the validation loss does not decrease by more than $0.0001$ for $5$ epochs.

Initially we utilise augmentations and sampling strategies found in the literature and point cloud segmentation implementations.
A standard and fixed set of augmentations are applied in all training sequences.
All points within a sample are first shuffled for random order so that any structure present is removed. 
The sample is then shifted such that it is centred and the mean across $x, y, z$ values is zero.
Each sample is then augmented with down sampling between 0\% and 90\%, rotation around the $z$ axis, and random noise of $0.5m$ is added to points.

We measure model performance using intersection over union (IOU) across the datasets. 
Where available we break down the performance on subsets of each dataset, e.g. for Swiss3DCities we evaluate performance on each city and density individually.
To evaluate the overall generalisation performance we take the mean of the IoU's over each class, then over the dataset's subsets and finally over each test dataset.
We define IoU as shown in equation \eqref{eq:IoU} where $TP_c$, $FN_c$, and $FP_c$ are the true positive, false negative, and false positive labels for each class $_c$.
\begin{equation}
    IoU_c = \frac{TP_c}{TP_c + FN_c + FP_c}
    \label{eq:IoU}
\end{equation}

\section{Point Cloud Sampling}\label{sec:sampling}
Aerial point cloud datasets are typically made up of large areas which are tiled for manageable file sizes.
These tiles often contain too much data to process through a model in a single pass.
Methods of sampling these tiles select a retrieve a set number of points.
This does not typically pose an issue on a single dataset since the variability of density is not as high.
In the situation where there is considerable variability in density of the source data the model must learn to generalise across this variable.

\subsection{Naive solution}
The naive approach takes the standard implementation of sampling from the larger point clouds using $N$ points around a single source point.
The resulting sample is a spherical encapsulation surrounding the source point.
We train on each individual training dataset and then on the combination of both. 
The results of the training and evaluation are shown in Table \ref{tab:baslinetests}.
The Overall dataset refers to the mean of each type of dataset/type, for example ``Overall LIDAR" refers to the mean performance of the model on all LIDAR test sets.

\begin{table}
    \centering
    \caption{Mean IoU across classes ground, building, and vegetation of segmentation models. Bold numbers signify highest performance on each test set.}
    \resizebox{0.45\textwidth}{!}{
        \begin{tabular}{c||c|c|c}
            \hline
                                    & \multicolumn{3}{c}{\textbf{Training set}}         \\ \hline \hline
            \textbf{Test Set}       & \textbf{DALES} & \textbf{SUM}           & \textbf{DALES-SUM}     \\ \hline \hline
            DALES                   & \textbf{0.949} & 0.654                  & 0.807                  \\ \hline
            SUM                     & 0.5            & \textbf{0.728}         & 0.623                  \\ \hline \hline
            ISPRS                   & \textbf{0.811} & 0.566                  & 0.713                  \\ \hline
            Sensat                  & 0.345          & 0.449                  & \textbf{0.548}         \\ \hline \hline
            H3D all                 & 0.471          & 0.444                  & \textbf{0.527}         \\ \hline \hline
            H3D LIDAR               & 0.503          & 0.460                  & \textbf{0.599}         \\ \hline 
            H3D Mesh                & \textbf{0.423} & 0.419                  & 0.419                  \\ \hline \hline
            H3D 2016 LIDAR          & \textbf{0.778} & 0.415                  & 0.656                  \\ \hline 
            H3D 2018 LIDAR          & 0.372          & 0.497                  & \textbf{0.556}         \\ \hline 
            H3D 2018 Mesh           & \textbf{0.422} & 0.416                  & 0.408                  \\ \hline 
            H3D 2019 LIDAR          & 0.358          & 0.469                  & \textbf{0.584}         \\ \hline 
            H3D 2019 Mesh           & 0.423          & 0.422                  & \textbf{0.429}         \\ \hline \hline
            Swiss3DCities All       & 0.5204         & 0.5246                 & \textbf{0.5546}        \\ \hline \hline
            Swiss3DCities Sparse    & \textbf{0.457} & 0.372                  & 0.389                  \\ \hline
            Swiss3DCities Medium    & 0.553          & 0.585                  & \textbf{0.612}         \\ \hline
            Swiss3DCities Zurich    & 0.509          & \textbf{0.57}          & 0.553                  \\ \hline
            Swiss3DCities Davos     & 0.533          & 0.556                  & \textbf{0.616}         \\ \hline
            Swiss3DCities Zug       & 0.55           & 0.54                   & \textbf{0.603}         \\ \hline \hline
            Overall LIDAR          & 0.598          & 0.473                  & \textbf{0.606}         \\ \hline
            Overall Photogrammetry & 0.429          & 0.464                  & \textbf{0.507}         \\ \hline
            Overall            & 0.526          & 0.475                  & \textbf{0.569}         \\ \hline
            Overall (STD)      & 0.173          & \textbf{0.069}         & 0.083                  \\ \hline
        \end{tabular}
    } 
    \label{tab:baslinetests}
\end{table}

We include the DALES and SUM performance evaluation in this experiment to show that each model performs well on the individually trained datasets. 
These values are not included in the overall metrics at the bottom of the table.
The performance of the DALES and SUM models on their test sets aligns with the IoU's seen in their respective papers for the classes included.
This indicates that the model is trained to a similar quality to those benchmarked in the papers and therefore the generalisation performance can be inferred.

Initially we observe that models trained on LIDAR data alone do not transfer well to photogrammetry data.
Although the performance is lower for models trained only on photogrammetry data, the results are much more consistent and predictable.
This could be due to the fact that typically photogrammetry compared to LIDAR contains much more noise and error.
Interestingly the DALES only trained model has the highest performance on H3D photogrammetry data.

It is unintuitive to see that the model performance on Swiss3DCities Davos, which contains a mountainous topography, appears to not be as challenging as the dense urban area of Zurich.
Although Davos and Zurich are both urban areas, the higher density of Zurich may be a more challenging task.
The H3D photogrammetry data appears to be more difficult for the model trained on the combined set.
We can also see that the ISPRS dataset is likely the simplest in the test set, consistently having the highest mean IoU for each model.

The results of training on DALES and SUM datasets individually both clearly show the advantage of the combined training set.
In most cases the model trained on the combined dataset has a higher mean IoU than either model trained on the individual datasets. 
This is a strong signal that the combination of the datasets causes the model to learn better features that are more general to all aerial point clouds.
Overall a 8.3\% and 19.7\% increase in IoU on DALES and SUM, respectively is observed by combining the training sets.
The positive generalisation performance results of this experiment motivate our further exploration.

\subsection{Constant radius}\label{sec:ConstantRadius}
We observed that there was a large disparity between the performance on the sparse and medium subsets of Swiss3D test set.
The point density differences between these sets is quite large and with the sampling method would result in a sample radius of between $18m$ and $109m$.
This large disparity in sample radius and variable distances between points may confuse the model and produce the observed poor results.
Constant radius sampling attempts to add consistency to the samples which are given to the model.

In these experiments the model will only have to identify larger structures which should remain in the similar overall scale with a limited density.
The maximum density can be calculated as: $d_{max} = N/(\pi r^2)$, where $N$ is the number of points and $r$ is the radius.
In the case of a sample radius of $145m$ the maximum density will be $~1 pt/m^2$.
All datasets in the test set have a density higher than $1 pt/m^2$ and hence will not have any padding or upsampling with a sample radius of $145m$.
In the case of a radius of $30m$ however the lower density point clouds will need to be upsampled.

To achieve the constant radius sampling we query the KD-tree for all points up to a set radius. 
This produces a variable number of points which cannot be batched to pass through the model.
To address the variable number of points we either randomly drop points or upsample the number of points until the desired amount is met.
Upsampling in this experiment is done by taking random points in the sample and duplicating them with noise.
Upsampled points are then ignored in the evaluation.

\begin{table}
    \centering
    \caption{Models trained on DALES and SUM dataset with constant radius sampling. Percentage change (PC) shown over the naive solution trained on DALES and SUM in Table \ref{tab:baslinetests}.}
    \resizebox{0.45\textwidth}{!}{
        \begin{tabular}{c||c|c|c|c|c|c}
            \hline
                                    & \multicolumn{6}{c}{\textbf{Training Radius}}  \\ \hline \hline
            \textbf{Test Set}       & \textbf{30}    & \textbf{40}    & \textbf{50}            & \textbf{70}    & \textbf{100}   & \textbf{145}   \\ \hline \hline
            ISPRS                   & 0.735          & 0.677          & 0.718                  & 0.691          & 0.738          & \textbf{0.776} \\ \hline
            Dublin                  & 0.577          & 0.681          & \textbf{0.73}          & 0.7            & 0.698          & 0.708          \\ \hline
            Sensat                  & 0.781          & \textbf{0.84}  & 0.839                  & 0.815          & 0.804          & 0.783          \\ \hline \hline
            H3D all                 & 0.553          & 0.555          & \textbf{0.566}         & 0.487          & 0.530          & 0.561          \\ \hline \hline
            H3D LIDAR               & \textbf{0.641} & 0.612          & 0.620                  & 0.541          & 0.600          & 0.627          \\ \hline
            H3D Mesh                & 0.421          & 0.470          & \textbf{0.485}         & 0.405          & 0.425          & 0.463          \\ \hline \hline
            H3D 2016 LIDAR          & 0.635          & 0.623          & 0.619                  & 0.571          & 0.642          & \textbf{0.652} \\ \hline
            H3D 2018 LIDAR          & \textbf{0.653} & 0.597          & 0.607                  & 0.498          & 0.521          & 0.592          \\ \hline
            H3D 2018 Mesh           & 0.426          & 0.472          & \textbf{0.488}         & 0.403          & 0.438          & 0.463          \\ \hline
            H3D 2019 LIDAR          & 0.635          & 0.617          & 0.635                  & 0.555          & \textbf{0.637} & 0.636          \\ \hline
            H3D 2019 Mesh           & 0.416          & 0.468          & \textbf{0.482}         & 0.406          & 0.412          & 0.463          \\ \hline \hline
            Swiss3DCities All       & 0.606          & 0.681          & \textbf{0.689}         & 0.683          & 0.661          & 0.627          \\ \hline \hline
            Swiss3DCities Sparse    & 0.548          & 0.500          & 0.579                  & \textbf{0.586} & 0.560          & 0.526          \\ \hline
            Swiss3DCities Medium    & 0.626          & \textbf{0.748} & 0.736                  & 0.725          & 0.711          & 0.666          \\ \hline
            Swiss3DCities Zurich    & 0.564          & \textbf{0.710}  & 0.69                   & 0.696          & 0.679          & 0.625         \\ \hline
            Swiss3DCities Davos     & 0.679          & 0.753          & \textbf{0.756}         & 0.724          & 0.687          & 0.669          \\ \hline
            Swiss3DCities Zug       & 0.615          & \textbf{0.693} & 0.684                  & 0.684          & 0.667          & 0.65           \\ \hline \hline
            Overall LIDAR          & 0.651          & 0.657          & 0.689                  & 0.644          & 0.679          & \textbf{0.704} \\ \hline
            Overall Photogrammetry & 0.603          & 0.664          & \textbf{0.671}         & 0.634          & 0.630          & 0.624          \\ \hline
            Overall            & 0.650          & 0.687          & \textbf{0.708}         & 0.675          & 0.686          & 0.691          \\ \hline
            Overall STD        & 0.101          & 0.101          & 0.098                  & 0.118          & 0.102          & \textbf{0.096} \\ \hline
            PC over naïve solution  & 114.23\%       & 120.62\%       & \textbf{124.41\%}      & 118.56\%       & 120.50\%       & 121.36\%       \\ \hline
        \end{tabular}
    }

    \label{tab:constantradius}
\end{table}

Through constraining the radius of the sample it was found that there was between a 0.05 and 0.11 (14.2\% and 24.4\%) increase in IoU over the DALES-SUM training set.
This boost in performance appears to substantially come from increased and more consistent IoU performance on photogrammetry data.
The performance of a $50m$ constant radius vs a naive DALES-SUM trained model increases by 13.7\% on LIDAR data and by 32.3\% on photogrammetry data.

If we compare these models with a model simply trained on SUM or DALES the improvement is quite significant.
Our results show an increase in performance of 35\% and 49\% on the DALES and SUM only training, respectively.

\subsection{Constant density}\label{sec:ConstantDensity}
In the previous experiment we observe that there is an advantage for model performance if samples are more consistent.
It is unclear if this is the case due to constant radius, more regular densities, or both.
Since we must have a fixed number of points for each model and a constant density, a constant radius is also needed.
In this experiment the radius is fixed to $30m$ and the density is constant for each model trained.
This is accomplished by changing the number of points the model processes in each pass.

\begin{table}    
    \centering
    \caption{Models trained on DALES and SUM dataset with constant 30m radius and variable density sampling. Percentage change (PC) shown over the naive solution trained on DALES and SUM in Table \ref{tab:baslinetests}.}
    \resizebox{0.45\textwidth}{!}{
        \begin{tabular}{c|cccccc}
            \hline
                                    & \multicolumn{6}{c}{\textbf{Training Density ($pts/m^2$)}}                                    \\ \hline \hline
            \textbf{Test Set}       & \textbf{1}      & \textbf{2}        & \textbf{4} & \textbf{8}   & \textbf{16}& \textbf{32}  \\ \hline \hline
            ISPRS                   & 0.732           & \textbf{0.782}    & 0.771    & 0.77           & 0.7      & 0.496          \\ \hline
            Dublin                  & 0.685           & 0.718             & 0.72     & \textbf{0.728} & 0.716    & 0.682          \\ \hline
            Sensat                  & 0.803           & 0.808             & 0.812    & \textbf{0.832} & 0.829    & 0.831          \\ \hline \hline
            H3D all                 & 0.585           & \textbf{0.586}    & 0.574    & 0.551          & 0.543    & 0.518          \\ \hline \hline
            H3D LIDAR               & 0.633           & \textbf{0.653}    & 0.641    & 0.636          & 0.619    & 0.582          \\ \hline
            H3D Mesh                & \textbf{0.513}  & 0.485             & 0.474    & 0.424          & 0.430    & 0.421          \\ \hline \hline
            H3D 2016 LIDAR          & 0.643           & \textbf{0.668}    & 0.66     & 0.656          & 0.603    & 0.549          \\ \hline
            H3D 2018 LIDAR          & 0.611           & \textbf{0.635}    & 0.622    & 0.617          & 0.621    & 0.566          \\ \hline
            H3D 2018 Mesh           & \textbf{0.502}  & 0.484             & 0.478    & 0.429          & 0.418    & 0.402          \\ \hline
            H3D 2019 LIDAR          & 0.645           & \textbf{0.657}    & 0.64     & 0.634          & 0.633    & 0.632          \\ \hline
            H3D 2019 Mesh           & \textbf{0.523}  & 0.485             & 0.47     & 0.418          & 0.442    & 0.44           \\ \hline \hline 
            Swiss3DCities All       & \textbf{0.711}  & 0.692             & 0.699    & 0.704          & 0.674    & 0.645          \\ \hline \hline
            Swiss3DCities Sparse    & \textbf{0.671}  & 0.621             & 0.613    & 0.608          & 0.460    & 0.311          \\ \hline
            Swiss3DCities Medium    & 0.727           & 0.726             & 0.735    & 0.745          & 0.745    & \textbf{0.746} \\ \hline
            Swiss3DCities Zurich    & 0.714           & 0.697             & 0.71     & \textbf{0.726} & 0.725    & \textbf{0.726} \\ \hline
            Swiss3DCities Davos     & \textbf{0.772}  & 0.731             & 0.74     & 0.738          & 0.738    & 0.739          \\ \hline
            Swiss3DCities Zug       & 0.671           & 0.684             & 0.695    & \textbf{0.702} & 0.701    & \textbf{0.702} \\ \hline \hline
            Overall LIDAR          & 0.683           & \textbf{0.718}    & 0.711    & 0.711          & 0.678    & 0.587          \\ \hline
            Overall Photogrammetry & \textbf{0.676}  & 0.661             & 0.662    & 0.653          & 0.644    & 0.632          \\ \hline
            Overall            & 0.703           & \textbf{0.717}    & 0.715    & \textbf{0.717}          & 0.692    & 0.634          \\ \hline
            Overall (STD)      & \textbf{0.079}  & 0.087             & 0.090    & 0.105          & 0.102    & 0.136          \\ \hline
            PC over naïve solution  & 123.49\%        & \textbf{125.93\%} & 125.58\% & 125.90\%       & 121.60\% & 111.39\%      \\ \hline
        \end{tabular}
    }

    \label{tab:constantradiusvardensity}
\end{table}

Training the model with constant density shows that the RandLA-Net model does not continuously improve in performance given higher density data.
This can be observed in the datasets which contain higher densities such as Swiss3DCitites Medium.
We can observe that the higher testing densities have a positive effect on performance but overall plateaus at an mean IoU of 0.746 in Swiss3DCities Medium.
The performance on each dataset which contains high density data appears to peak or plateau.

We observe a plateau in performance in high density dataset from around $8 pts/m^2$.
The overall performance is then degraded by subsets such as Swiss3DCities Sparse which contain low density data.
This is a product of the upsampling strategy utilised for our experiments.
In the case of the Swiss3DCities Sparse subset, a density as high as $32 pts/m^2$ will cause each point to be replicated almost $20$ times.
This clearly has detrimental effects shown by the decrease in IoU as density increases.

The overall performance of the model on the test sets peaks at $2 pts/m^2$ and begins to decrease at $16 pts/m^2$.
We observe only a 2\% decrease in performance from $1 pts/m^2$ indicating that the model is well equipped to handle lower density data.
The plateau and decrease as test density increases indicates that the model doesn't take advantage of the high density datasets as would be assumed.

\subsection{Constant radius and density}
This experiment attempts to isolate all factors except for the change in radius to observe if there is a particular radius which performs best.
To test this we fix the density of the samples to be $1 pts/m^2$. 
We achieve this by fixing the density and changing the number of points for each sample.
Fixing this value also decreases the number of points the model handles with each sample.

\begin{table}
    \centering
    \caption{Models trained on DALES and SUM dataset with constant radius and density sampling. Percentage change (PC) shown over the naive solution trained on DALES and SUM in Table \ref{tab:baslinetests}.}
    \resizebox{0.45\textwidth}{!}{
        \begin{tabular}{c||c|c|c|c|c}
        \hline
                                & \multicolumn{5}{c}{\textbf{Training (N/radius)}}  \\ \hline \hline
        \textbf{Test Set}       & $2^{12}/36$              & $2^{13}/51$              & $2^{14}/72$              & $2^{15}/102$              & $2^{16}/145$              \\ \hline \hline
        ISPRS                   & 0.732                    & 0.743                    & 0.749                    & 0.757                     & \textbf{0.776}            \\ \hline
        Dublin                  & 0.685                    & 0.68                     & 0.69                     & \textbf{0.741}            & 0.708                     \\ \hline
        Sensat                  & 0.803                    & 0.806                    & \textbf{0.808}           & 0.806                     & 0.783                     \\ \hline \hline
        H3D all                 & \textbf{0.585}           & 0.540                    & 0.529                    & 0.552                     & 0.561                     \\ \hline \hline 
        H3D LIDAR               & 0.633                    & 0.619                    & 0.569                    & \textbf{0.644}            & 0.627                     \\ \hline
        H3D Mesh                & \textbf{0.513}           & 0.414                    & 0.469                    & 0.414                     & 0.463                     \\ \hline \hline
        H3D 2016 LIDAR          & 0.643                    & 0.618                    & 0.55                     & 0.646                     & \textbf{0.652}            \\ \hline
        H3D 2018 LIDAR          & 0.611                    & 0.593                    & 0.564                    & \textbf{0.623}            & 0.592                     \\ \hline
        H3D 2018 Mesh           & \textbf{0.502}           & 0.426                    & 0.462                    & 0.403                     & 0.463                     \\ \hline
        H3D 2019 LIDAR          & 0.645                    & 0.617                    & 0.593                    & \textbf{0.663}            & 0.636                     \\ \hline
        H3D 2019 Mesh           & \textbf{0.523}           & 0.444                    & 0.475                    & 0.425                     & 0.463                     \\ \hline \hline
        Swiss3DCities All       & \textbf{0.711}           & 0.696                    & 0.683                    & 0.663                     & 0.627                     \\ \hline \hline
        Swiss3DCities Sparse    & \textbf{0.671}           & 0.652                    & 0.624                    & 0.589                     & 0.526                     \\ \hline
        Swiss3DCities Medium    & \textbf{0.727}           & 0.724                    & 0.717                    & 0.697                     & 0.666                     \\ \hline
        Swiss3DCities Zurich    & \textbf{0.714}           & 0.691                    & 0.687                    & 0.662                     & 0.625                     \\ \hline
        Swiss3DCities Davos     & \textbf{0.772}           & 0.738                    & 0.733                    & 0.704                     & 0.669                     \\ \hline
        Swiss3DCities Zug       & 0.671                    & \textbf{0.674}           & 0.656                    & 0.663                     & 0.65                      \\ \hline \hline
        Overall LIDAR          & 0.683                    & 0.681                    & 0.669                    & \textbf{0.714}            & 0.704                     \\ \hline
        Overall Photogrammetry & \textbf{0.676}           & 0.639                    & 0.653                    & 0.628                     & 0.624                     \\ \hline
        Overall            & 0.703                    & 0.693                    & 0.692                    & \textbf{0.704}            & 0.691                     \\ \hline
        Overall (STD)      & \textbf{0.079}          & 0.099                    & 0.104                    & 0.099                     & 0.096                     \\ \hline 
        PC over naïve solution  & 123.49\%                 & 121.68\%                 & 121.49\%                 & \textbf{123.60\%}         & 121.36\%                 \\ \hline
        \end{tabular}
    }
    
    \label{tab:constantradandden}
\end{table}

We observe that performance of the models do not change substantially across the different radii.
In fact we observe that the overall performance is very consistent across the variable radii in these tests.
This performance consistency indicates that it is the constant density which has the largest impact in performance and not a particular radius.

In this experiment the model is trained on between $2^{12}$ and $2^{16}$ points.
Comparing the constant radius of $30m$ with $2^{16}$ points performance of overall IoU of $0.650$ from table \ref{tab:constantradius} and the $36m$ radius with only $2^{12}$ points at $0.703$.
The results show that there is a considerable generalisation performance increase of 8\% with constant density between all samples.

\section{Discussion}
In this section we briefly discuss the overall observations we have found through our experimentation. 
This includes observations of performance on the combined test set as well as the presented sampling strategies.
\begin{table}
    \centering
    \caption{Meta analysis of the combined test set showing some key statistics over Tables \ref{tab:constantradius},\ref{tab:constantradiusvardensity}, and \ref{tab:constantradandden}.}
    \resizebox{0.45\textwidth}{!}{
        \begin{tabular}{c||c|c|c|c}
            \hline
            \textbf{Test Set}       & \textbf{Mean IoU} & \textbf{STD}   & \textbf{Maximum} & \textbf{Minimum} \\ \hline \hline
            ISPRS                   & 0.726    & 0.067 & 0.782   & 0.496   \\ \hline
            Dublin                  & 0.697    & 0.036 & 0.741   & 0.577   \\ \hline
            Sensat                  & 0.811    & 0.019 & 0.840   & 0.781   \\ \hline \hline
            H3D all                 & 0.551    & 0.026 & 0.586   & 0.487   \\ \hline \hline
            H3D LIDAR               & 0.618    & 0.029 & 0.653   & 0.541   \\ \hline
            H3D Mesh                & 0.452    & 0.035 & 0.513   & 0.405   \\ \hline \hline
            H3D 2016 LIDAR          & 0.625    & 0.037 & 0.668   & 0.549   \\ \hline
            H3D 2018 LIDAR          & 0.595    & 0.040 & 0.653   & 0.498   \\ \hline
            H3D 2018 Mesh           & 0.451    & 0.035 & 0.502   & 0.402   \\ \hline
            H3D 2019 LIDAR          & 0.630    & 0.025 & 0.663   & 0.555   \\ \hline
            H3D 2019 Mesh           & 0.456    & 0.036 & 0.523   & 0.406   \\ \hline \hline
            Swiss3DCities All       & 0.674    & 0.031 & 0.711   & 0.606   \\ \hline \hline
            Swiss3DCities Sparse    & 0.567    & 0.088 & 0.671   & 0.311   \\ \hline
            Swiss3DCities Medium    & 0.716    & 0.034 & 0.748   & 0.626   \\ \hline
            Swiss3DCities Zurich    & 0.685    & 0.044 & 0.726   & 0.564   \\ \hline
            Swiss3DCities Davos     & 0.726    & 0.033 & 0.772   & 0.669   \\ \hline
            Swiss3DCities Zug       & 0.674    & 0.023 & 0.702   & 0.615   \\ \hline \hline
            Overall LIDAR          & 0.680    & 0.033 & 0.718   & 0.587   \\ \hline
            Overall Photogrammetry & 0.646    & 0.021 & 0.676   & 0.603   \\ \hline
            Overall            & 0.692    & 0.022 & 0.717   & 0.634   \\ \hline
            Overall (STD)      & 0.100    & 0.013 & 0.136   & 0.079   \\ \hline
            PC over naïve solution  & 121.5\%  & 3.9\% & 125.9\% & 111.4\%  \\ \hline
        \end{tabular}
    }
    \label{tab:metaanalysis}
\end{table}

\subsection{Meta analysis of test set}
Through combining the currently available aerial point cloud segmentation datasets we have presented a challenging generalisation baseline.
This baseline allows us to draw conclusions on model performance in key characteristics such as sensor type, data quality, and content variability.
Performance across these characteristics is important for developing models which can perform reliably and predictably on variable data.

Table \ref{tab:metaanalysis} shows an overview of performance of the model on each of the test datasets and their sub-tasks.
Overall we observe that the model performs worse on photogrammetry data than LIDAR but this is mainly due to the poor performance on the H3D Mesh data.

Is can be observed that the overall performance on the H3D mesh data is much lower than the other photogrammetry datasets of Swiss3DCities and Sensat.
We believe this to be the case due to the poorer quality point cloud reconstruction as shown in figure \ref{fig:h3dmeshLIDARcomparison}.
Achieving higher performance on this subset could give confidence that the model will transfer well to noisier photogrammetry data.
We have found that H3D Mesh data is the most challenging dataset while the models perform highest on the Sensat dataset.

We also observe that our sampling strategy has fairly consistent performance across each time period of the H3D dataset despite varying qualities and densities of the point clouds.
The highest variance in performance belongs to the Swiss3DCities Sparse subset which suffers from the upsampling strategy discussed in section \ref{sec:ConstantRadius}.
Despite matching the density differences between Swiss3DCities Sparse and Medium by down sampling both to $1pt/m^2$ we still observe discrepancy in performance. 
Our results show that addressing the upsampling issues at the model level will result in a considerable improvement in generalisation performance.

Throughout the experimentation the largest impact in performance is from model confusion between buildings and vegetation.
In the case of photogrammetry, vegetation will resemble a shell while LIDAR presents as a mass of points. 
This discrepancy is most prevalent in a model trained on LIDAR which struggles to perform on photogrammetry data.

\subsection{Sampling strategy observations}
We have observed that the sampling strategy used has a large impact on the performance of a model and its ability to generalise across multiple datasets.
We first observed that a constant radius sampling strategy improves the results and motivated our experiments to test which attributes effected the performance the most.
In our density experimentation from Section \ref{sec:ConstantDensity} we observe that overall performance peaks with densities lower than $16pts/m^2$ but increases and plateaus if the dataset has higher density.
This leads us to believe that a model which handles these variable densities directly and intentionally can maintain a much higher generalisation performance.

We have observed that overall a single sample radius size did not improve generalisation performance substantially within our tested ranges.
This was contrary to our belief that with wider context for each point, the model could be more certain of it's class.
It seems that a considerable challenge is the model's ability to represent local and overall features and hence each sample must contain this information.

\subsection{Future work}
In this paper we have not trained utilising the RGB or intensity channels of the datasets.
Despite this we believe that our method will increase the performance with the inclusion of the additional data.
An extension of this work would focus on utilising any additional channels included in the data.

Expanding these findings to multiple models, especially of the varying methods presented in section \ref{sec:segmentationmethods}, will be important.
We do believe that any methods which utilise similar sampling strategies can benefit from consistent samples.

\section{Conclusion}
In this paper we have combined available aerial point cloud datasets to evaluate a segmentation model's generalisation performance.
This includes reasoning about which datasets to use as training/validation/testing, how we combine classes to larger categories, and important characteristics.
This combination allows researchers to evaluate performance across different sensor types, contents of datasets, and variability in quality.
We have observed that there is a challenge in handling these variables.
We have shown that a simple sampling strategy can increase performance of the model substantially over a naive approach.
We hope that this paper motivates researchers to utilise this combination of datasets as a generalisation benchmark to compare performance to.
We would also like to see datasets increase consistency in classes and formatting to aid in additions to our generalisation evaluation test set in the future.

\section*{Acknowledgements}
This research was supported by Australian Government Research Training Program (RTP) Scholarships awarded to Matthew Howe.
This paper was developed in collaboration with Defence Industry including Lockheed Martin Australia, the Australian Institute for Machine Learning, and the University of Adelaide.

{\small
\bibliographystyle{ieee_fullname}
\bibliography{references}
}

\end{document}